\documentclass[runningheads]{llncs}
\usepackage{graphicx}
\usepackage{amsmath,amssymb} 
\usepackage{color}

\usepackage{cite}
\usepackage{graphicx}
\usepackage{wrapfig}
\usepackage{subfig}
\usepackage{parskip}
\usepackage{color}
\usepackage{xspace}
\usepackage{overpic}
\usepackage{subfig}
\usepackage{wrapfig}
\usepackage{enumitem}
\usepackage{mathrsfs}
\usepackage{helvet}
\usepackage{times}
\usepackage{textcomp}
\usepackage{multirow}

\definecolor{turquoise}{cmyk}{0.65,0,0.1,0.1}
\definecolor{purple}{rgb}{0.65,0,0.65}
\definecolor{dark_green}{rgb}{0, 0.5, 0}
\definecolor{orange}{rgb}{0.8, 0.6, 0.2}
\definecolor{red}{rgb}{0.8, 0.2, 0.2}
\definecolor{brown}{rgb}{0.5, 0.16, 0.16}

\newcommand{\etal}{et al.}

\usepackage{lipsum}

\usepackage[width=122mm,left=12mm,paperwidth=146mm,height=193mm,top=12mm,paperheight=217mm]{geometry}
\begin{document}
\pagestyle{headings}
\mainmatter
\def\ECCV18SubNumber{1958}  

\title{Specular-to-Diffuse Translation \\ for Multi-View Reconstruction} 

\titlerunning{Specular-to-Diffuse Translation for Multi-View Reconstruction}

\authorrunning{S. Wu, H. Huang, T. Portenier, M. Sela, D. Cohen-Or, R. Kimmel and M. Zwicker}

\author{
Shihao Wu$^{1}$%
\hspace{18pt}
Hui Huang$^{2}$  \thanks{Corresponding author: Hui Huang (hhzhiyan@gmail.com)}
\hspace{18pt}
Tiziano Portenier$^{1}$ 
\hspace{18pt}
Matan Sela$^{3}$ 
\\
\hspace{18pt}
Daniel Cohen-Or$^{4}$ 
\hspace{18pt}
Ron Kimmel$^{3}$ %
\hspace{18pt}
Matthias Zwicker$^{5}$
\\
}

\institute{
$^{1}$University of Bern %
\hspace{8pt}
$^{2}$Shenzhen University
\hspace{8pt}
$^{3}$Technion - Israel Institute of Technology
\hspace{8pt}
$^{4}$Tel-Aviv University %
\hspace{8pt}
$^{5}$University of Maryland %
}

\maketitle

\begin{abstract}

Most multi-view 3D reconstruction algorithms, especially when shape-from-shading cues are used, assume that object appearance is predominantly diffuse. To alleviate this restriction, we introduce S2Dnet, a generative adversarial network for transferring multiple views of objects with specular reflection into diffuse ones, so that  multi-view reconstruction methods can be applied more effectively. Our network extends unsupervised image-to-image translation to multi-view ``specular to diffuse" translation. To preserve object appearance across multiple views, we introduce a Multi-View Coherence loss (MVC) that evaluates the similarity and faithfulness of local patches after the view-transformation. Our MVC loss ensures that the similarity of local correspondences among multi-view images is preserved under the image-to-image translation. As a result, our network yields significantly better results than several single-view baseline techniques. In addition, we carefully design and generate a large synthetic training data set using physically-based rendering. During testing, our network takes only the raw glossy images as input, without extra information such as segmentation masks or lighting estimation. Results demonstrate that multi-view reconstruction can be significantly improved using the images filtered by our network. We also show promising performance on real world training and testing data.

\end{abstract}

\section{Introduction}
\label{sec:intro}

Three-dimensional reconstruction from multi-view images is a long standing problem in computer vision. State-of-the-art shape-from-shading techniques achieve impressive results \cite{Langguth2016, maier2017intrinsic3d}. These techniques, however, make rather strong assumptions about the data, mainly that target objects are predominantly diffuse with almost no specular reflectance. Multi-view reconstruction of glossy surfaces is a challenging problem, which has been addressed by adding specialized hardware (e.g., coded pattern projection \cite{tarini20053d} and two-layer LCD \cite{tin20163d}), imposing surface constraints \cite{ikeuchi1981determining, savarese2001local}, or making use of additional information like silhouettes and environment maps \cite{godard2015multi}, or the Blinn-Phong model \cite{khanian2017photometric}. 

\begin{figure*}[t!]
\centering
	{\includegraphics[width=.99\linewidth]{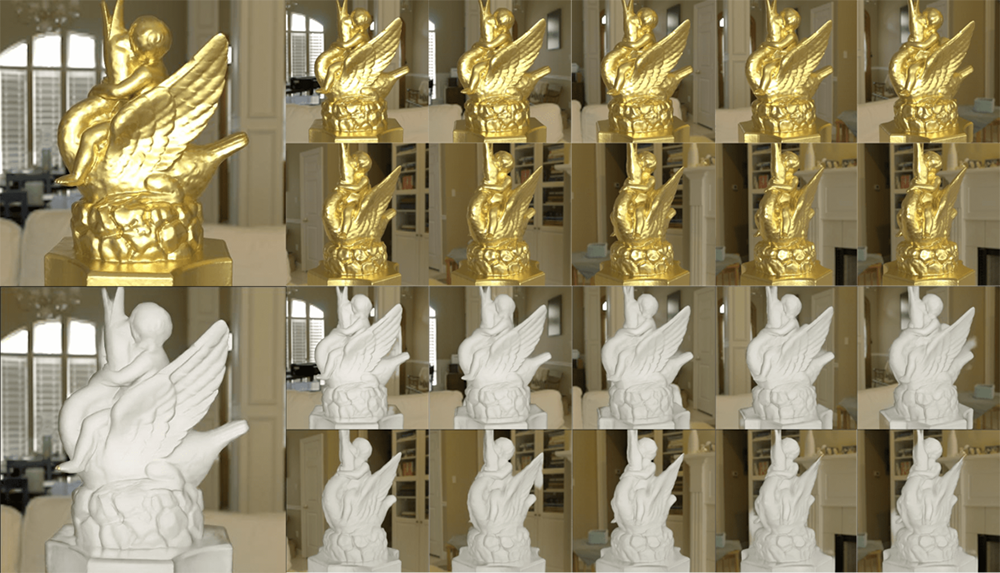}}

	\caption{Specular-to-diffuse translation of multi-view images. We show eleven views of a glossy object (top), and the specular-free images generated by our network (bottom).}
\label{fig_teaser}
\end{figure*}

In this paper, we present a generative adversarial neural network (GAN) that translates multi-view images of objects with specular reflection to diffuse ones. The network aims to generate a specular-free surface, which then can be reconstructed by a standard multi-view reconstruction technique as shown in Figure~\ref{fig_teaser}. We name our translation network, S2Dnet, for Specular-to-Diffuse. Our approach is inspired by recent GAN-based image translation methods, like pix2pix \cite{isola2017image} or cycleGAN \cite{zhu2017unpaired}, that can transform an image from one domain to another. Such techniques, however, are not designed for multi-view image translation. Directly applying these translation techniques to individual views is prone to reconstruction artifacts due to the lack of coherence among the transformed images. Hence, instead of using single views, our network considers a triplet of nearby views as input. These triplets allow learning the mutual information of neighboring views. More specifically, we introduce a global-local discriminator and a perceptual correspondence loss that evaluate the multi-view coherency of local corresponding image patches. Experiments show that our method outperforms baseline image translation methods. 

Another obstacle of applying image translation techniques to specularity removal is the lack of good training data. It is rather impractical to take enough paired or even unpaired photos to successfully train a deep network. Inspired by the recent works of simulating training data by physically-based rendering \cite{zhang2017physically, movshovitz2016useful, shi2017learning, meka2018live} and domain adaptation \cite{hoffman2017cycada, benaim2017one, kim2017learning, kang2018deep}, we present a fine-tuned process for generating training data, then adapting it to real world data. Instead of using Shapenet \cite{shapenet2015}, we develop a new training dataset that includes models with richer geometric details, which allows us to apply our method to complex real-world data. Both quantitative and qualitative evaluations demonstrate that the performance of multi-view reconstruction can be significantly improved using the images filtered by our network. We show also the performance of adapting our network on real world training and testing data with some promising results.

%
%

\section{Related work}
\label{sec:related}


{\bf Specular Object Reconstruction.}
%
Image based 3D reconstruction has been widely used for AR/VR applications, and the reconstruction speed and quality have been improved dramatically in recent years. However, most photometric stereo methods are based on the assumption that the object surface is diffuse, that is, the appearance of the object is view independent. Such assumptions, however, are not valid for glossy or specular objects in uncontrolled environments. It is well known that modeling the specularity is difficult as the specular effects are largely caused by the complicated global illumination that is usually unknown. For example, Godard et al. \cite{godard2015multi} first reconstruct a rough model by silhouette and then refine it using the specified environment map. Their method can reconstruct high quality specular surfaces from HDR images with extra information, such as silhouette and environment map. 

In contrast, our method requires only the multi-view images as input. Researchers have proposed sophisticated equipment, such as a setup with two-layer LCDs to encode the directions of the emitted light field \cite{tin20163d}, taking advantages of the IR images recorded by RGB-D scanners \cite{or2016real, or2015rgbd} or casting coded patterns onto mirror-like objects \cite{tarini20053d}. While such techniques can effectively handle challenging non-diffuse effects, they require additional hardware and user expertise. Another way to tackle this problem is by introducing additional assumptions, such as surface constraints \cite{ikeuchi1981determining, savarese2001local}, the Blinn-Phong model \cite{khanian2017photometric}, and shape-from-specularity \cite{chen2006mesostructure}. These methods can also benefit from our network that outputs diffuse images, where strong specularities are removed from uncontrolled illumination. Please refer to 
\cite{ihrke2010transparent} for a survey on specular object reconstruction.

{\bf GAN-based Image-to-Image Translation.}
%
We are inspired by the latest success of learning based image-to-image translation methods, such as ConditionalGAN \cite{isola2017image}, cycleGAN \cite{zhu2017unpaired}, \cite{yi2017dualgan} dualGAN, and discoGAN \cite{kim2017learning}. The remarkable capacity of Generative Adversarial Networks (GANs) \cite{goodfellow2014generative} in modeling data distributions allows these methods to transform images from one domain to another with relatively small amounts of training data, while preserving the intrinsic structure of original images faithfully. With improved multi-scale training techniques, such as Progressive GAN \cite{karras2017progressive} and pix2pixHD \cite{wang2017high}, image-to-image translation can be performed at mega pixel resolutions and achieve results of stunning visual quality. 

Recently, modified image-to-image translation architectures have been successfully applied to ill-posed or underconstrained vision tasks, including face frontal view synthesis \cite{huang2017beyond}, facial geometry reconstruction \cite{richardson20163d, sela2017unrestricted, richardson2017learning, sengupta2017sfsnet}, raindrop removal \cite{qian2017attentive}, or shadow removal \cite{wang2017stacked}. These applications motivate us to develop a glossiness removal method based on GANs to facilitate multi-view 3D reconstruction of non-diffuse objects.

{\bf Learning-based Multi-View 3D Reconstruction.}
%
Learning surface reconstruction from multi-view images end-to-end has been an active research direction recently \cite{tatarchenko2016multi, choy20163d, lin2017learning, tulsiani2017multi}. Wu et al. \cite{Wu2016} and Gwak et al. \cite{gwak2017weakly} use GANs to learn the latent space of shapes and apply it to single image 3D reconstruction. 3D-R2N2 \cite{choy20163d} designs a recurrent network for unified single and multi-view reconstruction. Image2Mesh \cite{pontes2017image2mesh} learns parameters of free-form-deformation of a base model. Nonetheless, in general, the reconstruction quality of these methods cannot really surpass that of traditional approaches that exploit multiple-view geometry and heavily engineered photometric stereo pipelines. To take the local image feature coherence into account, we focus on removing the specular effect on the image level and resort to the power of multi-view reconstruction as a post-processing and also a production step. 

On the other hand, there are works, closer to ours, that focus on applying deep learning on subparts of the stereo reconstruction pipeline, such as depth and pose estimation \cite{zhou2017unsupervised}, feature point detection and description \cite{yi2016lift, detone2017superpoint}, semantic segmentation \cite{ma2017multi}, and bundle adjustment \cite{zhu2017object, zhu2017semantic}. These methods still impose the Lambertian assumption for objects or scenes, where our method can serve as a preprocessing step to deal with glossiness.

{\bf Learning-based Intrinsic Image Decomposition.}
%
Our method is also loosely related to some recent works on learning intrinsic image decomposition. These methods include training a CNN to reconstruct rendering parameters, e.g., material \cite{liu2017material, yu2017pvnn}, reflectance maps \cite{rematas2016deep}, illumination \cite{georgoulis2016delight}, or some combination of those components \cite{shi2017learning, georgoulis2017around, liu2017material}. These methods are often trained on synthetic data and are usually applied to the re-rendering of single images. Our method shares certain similarity with these methods. However, our goal is not to recover intrinsic images with albedos. Disregarding albedo, we aim for output images with a consistent appearance across the entire training set that reflects the structure of the object.

\section{Multi-view Specular-to-Diffuse GAN}

\begin{figure*}[t!]
\centering
{\includegraphics[width=.99\linewidth]{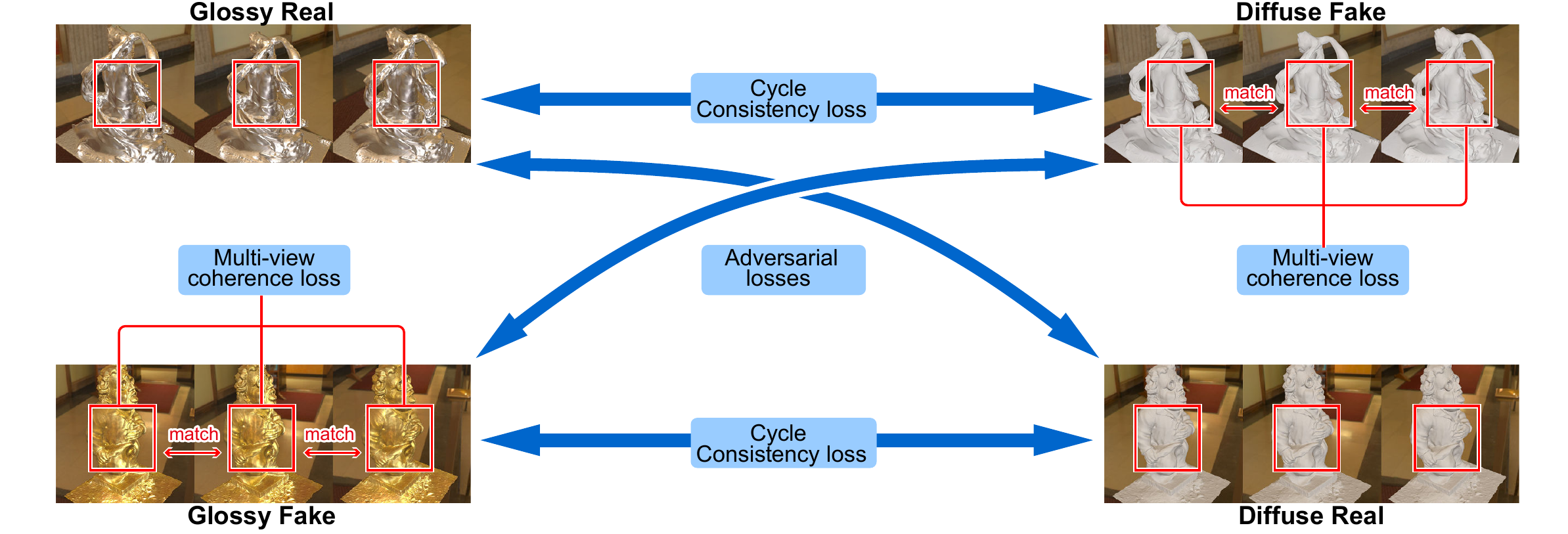}}
\caption{Overview of S2Dnet. Two generators and two discriminators are trained simultaneously to learn cross-domain translations between the glossy and the diffuse domain. In each training iteration, the model randomly picks and forwards a real glossy and diffuse image sequence, computes the loss functions and updates the model parameters.}
\label{fig_overview}
\end{figure*}
In this section, we introduce S2Dnet, a conditional GAN that translates multi-view images of highly specular scenes into corresponding diffuse images. The input to our model is a multi-view sequence of a glossy scene without any additional input such as segmentation masks, camera parameters, or light probes. This enables our model to process real-world data, where such additional information is not readily available. The output of our model directly serves as input to state-of-the-art photometric stereo pipelines, resulting in improved 3D reconstruction without additional effort. Figure~\ref{fig_overview} shows a visualization of the proposed model. We discuss the training data, one of our major contributions, in Section~\ref{sec:training_data}. In Section~\ref{sec:coherence} we introduce the concept of inter-view coherence that enables our model to process multiple views of a scene in a consistent manner, which is important in the context of multi-view reconstruction. Then, we outline in Section~\ref{sec:training_procedure} the overall end-to-end training procedure. Implementation details are discussed in Section~\ref{sec:impl_details}. Upon publication we will release both our data (synthetic and real) and the proposed model to foster further work.

\begin{figure*}[t!]
\centering
{\includegraphics[width=.99\linewidth]{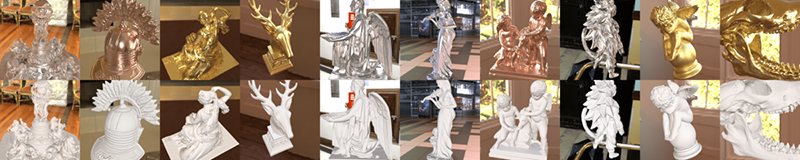}}
\caption{Gallery of our synthetically rendered specular-to-diffuse training data.}
\label{fig_gallery}
\end{figure*}

\subsection{Training Data}
\label{sec:training_data}
To train our model to translate multi-view glossy images to diffuse correspondents, we need appropriate data for both domains, i.e., glossy source domain images as inputs, and diffuse images as the target domain. Yi \etal \cite{yi2017dualgan} propose a MATERIAL dataset consisting of unlabeled data grouped in different material classes, such as plastic, fabric, metal, and leather, and they train GANs to perform material transfer. However, the MATERIAL dataset does not contain multi-view images and thus is not suited for our application. Moreover, the dataset is rather small and we expect our deep model to require a larger amount of training data. Hence, we propose a novel synthetic dataset consisting of multi-view images, which is both sufficiently large to train deep networks and complex to generalize to real-world objects. For this purpose, we collect and align 91 watertight and noise-free geometric models featuring rich geometric details from SketchFab (Figure~\ref{fig_gallery}). We exclude three models for testing and use the remaining 88 models for training. To obtain a dataset that generalizes well to real-world images, we use PBRT, a physically based renderer \cite{pbrt} to render these geometric models in various environments with a wide variety of glossy materials applied to form our source domain. Next, we render the target domain images by applying a Lambertian material to our geometric models.

Our experiments show that the choice of the rendering parameters has a strong impact on the translation performance. On one hand, making the two domains more similar by choosing similar materials for both domains improves the translation quality on synthetic data. Moreover, simple environments, such as a constant ground plane, also increase the quality on synthetic data. On the other hand, such simplifications cause the model to overfit and prevent generalization to real-world data. Hence, a main goal of our dataset is to provide enough complexity to allow generalization to real data.
%
%
%
To achieve realistic illumination, we randomly sample one of 20 different HDR indoor environment maps and randomly rotate it for each scene. In addition, we orient a directional light source pointing from the camera approximately towards the center of the scene and position two additional light sources above the scene. The intensities, positions, and directions of these additional light sources are randomly jittered. This setup guarantees a rather even, but still random illumination. To render the source domain images, we applied the various metal materials defined in PBRT, including copper, silver, and gold. Material roughness and index of refraction are randomly sampled to cover a large variety of glossy materials. We randomly sample camera positions on the upper hemisphere around the scene pointing towards the center of the scene. To obtain multi-view data, we always sample 5 close-by, consecutive camera positions in clock-wise order while keeping the scene parameters fixed to mimic the common procedure of taking photos for stereo reconstruction. Since we collect 5 images of the same scene and the input to our network consists of 3 views, we obtain 3 training samples per scene. All rendered images are of $512\times512$ resolution, which is the limit for our GPU memory. However, it is likely that higher resolutions would further improve the reconstruction quality. Finally, we render the exact same images again with a white, Lambertian material, i.e., the mapping from the source to the target domain is bijective. The proposed procedure results in a training dataset of more than 647k images, i.e., more than 320k images per domain. For testing, we rendered 2k sequences of images, each consisting of 50 images. All qualitative results on synthetic data shown in this paper belong to this test set.

\subsection{Inter-view Coherence}
\label{sec:coherence}
Multi-view reconstruction algorithms leverage corresponding features in different views to accurately estimate the 3D geometry. Therefore, we cannot expect good reconstruction quality if the glossy images in a multi-view sequence are translated independently using standard image translation methods, e.g., \cite{zhu2017unpaired, isola2017image}. This will introduce inconsistencies along the different views, and thus cause artifacts in the subsequent reconstruction. We therefore propose a novel model that enforces inter-view coherence by processing multiple views simultaneously. Our approach consists of a global and local consistency constraint: the global constraint is implemented using an appropriate network architecture, and the local consistency is enforced using a novel loss function.

\paragraph{\bf{Global Inter-view Coherence.}}
A straightforward idea to incorporate multiple views is to stack them pixel-by-pixel before feeding them to the network. We found that this does not lead to strong enough constraints, since the network can still learn independent filter weights for the different views. This results in blurry translations, especially if corresponding pixels in different views are not aligned, which is typically the case. 
Instead, we concatenate the different views along the spatial axis before feeding them to the network. This solution, although simple, enforces the network to use the same filter weights for all views, and thus effectively avoids 
inconsistencies on a global scale.

\paragraph{\bf{Local Inter-view Coherence.}}
\begin{figure*}[t!]
\centering
{\includegraphics[width=.49\linewidth]{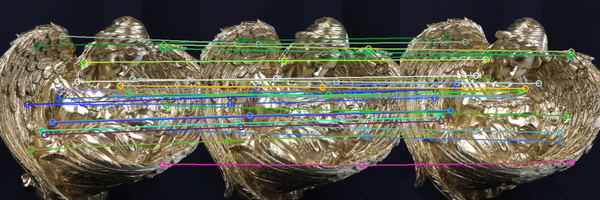}}
{\includegraphics[width=.49\linewidth]{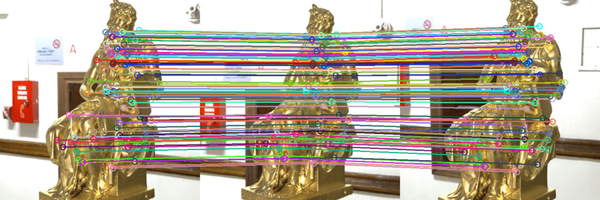}}
\caption{Two examples of the SIFT correspondences pre-computed for our training.}
\label{sift}
\end{figure*}
Incorporating loss functions based on local image patches has been successfully applied to generative adversarial models, such as image completion \cite{iizuka2017globally} or texture synthesis \cite{xian2017texturegan}. However, comparing image patches at random locations is not meaningful in a multi-view setup for stereo reconstruction. Instead, we encourage the network to maintain feature point correspondences in the input sequence, i.e., inter-view correspondences should be invariant to the translation. Since the subsequent reconstruction pipeline relies on such correspondences, maintaining them during translation should improve reconstruction quality. To achieve this, we first extract SIFT feature correspondences for all training images. For each training sequence consisting of three views, we compute corresponding feature points between the different views in the source domain; see Figure~\ref{sift} for two examples. During training, we encourage the network output at the SIFT feature locations to be similar along the views using a perceptual loss in VGG feature space \cite{gatys2016image,wang2017high, wang2017perceptual, vansteenkiste2017taming}. The key idea is to measure both high- and low-level similarity of two images by considering their feature activations in a deep CNN like VGG. We adopt this idea to keep local image patches around corresponding SIFT features perceptually similar in the translated output. The perceptual loss in VGG feature space is defined as:
\begin{equation}
\mathcal{L}_{VGG} (x, \hat x) = \sum_{i=1}^{N} \frac{1}{M_i} \|F^{(i)}(x) - F^{(i)}(\hat x)\|_1,
\end{equation}
where $F^{(i)}$ denotes the $i$-th layer in the VGG network consisting of $M_i$ elements. 
Now consider a glossy input sequence consisting of three images $X_1, X_2, X_3$, and the corresponding diffuse sequence $\tilde{X_1}, \tilde{X_2}, \tilde{X_3}$ produced by our model. A SIFT correspondence for this sequence consists of three image coordinates $p_1, p_2, p_3$, one in each glossy image, and all three pixels at the corresponding coordinates represent the same feature. We then extract local image patches $\tilde{x_i}$ centered at $p_i$ from $\tilde{X_i}$, and define the perceptual correspondence loss as:
\begin{equation}
\label{equation:correspondence}
\mathcal{L}_{corr}(\tilde{X_1}, \tilde{X_2}, \tilde{X_3}) = \mathcal{L}_{VGG}(\tilde{x_1}, \tilde{x_2}) + \mathcal{L}_{VGG}(\tilde{x_2}, \tilde{x_3}) + \mathcal{L}_{VGG}(\tilde{x_1}, \tilde{x_3}).
\end{equation}

\subsection{Training Procedure}
\label{sec:training_procedure}
Given two sets of data samples from two domains, a source domain $A$ and a target domain $B$, the goal of image translation is to find a mapping $T$ that transforms data points $X_i \in A$ to $B$ such that $T(X_i) = \tilde{X_i} \in B$, while the intrinsic structure of $X_i$ should be preserved under $T$. Training GANs has been proven to produce astonishing results on this task, both in supervised settings where the data of the two domains are paired \cite{isola2017image}, and in unsupervised cases using unpaired data \cite{zhu2017unpaired}. In our experiments, we observed that both approaches (ConditionalGAN \cite{isola2017image} and cycleGAN \cite{zhu2017unpaired}) perform similarly well on our dataset. However, while paired training data might be readily available for synthetic data, paired real-world data is difficult to obtain. Therefore we come up with a design for unsupervised learning that can easily be fine-tuned on unpaired real-world data.

\paragraph{\bf{Cycle-consistency Loss.}}
Similar to CycleGAN~\cite{zhu2017unpaired}, we learn the mapping between domain $A$ and $B$ with two translators $G_B : A \rightarrow B$ and $G_A : B \rightarrow A$ that are trained simultaneously. The key idea is to train with cycle-consistency loss, i.e., to enforce that $G_A(G_B(X)) \approx X$ and $G_B(G_A(Y)) \approx Y$, where $X \in A$ and $Y \in B$. This cycle-consistency loss guarantees that data points preserve their intrinsic structure under the learned mapping. Formally, the cycle-consistency loss is defined as:
\begin{equation}
\mathcal{L}_{cyc}(X, Y) = \|G_A(G_B(X)) - X\|_1 + \|G_B(G_A(Y)) - Y\|_1.
\end{equation}

\paragraph{\bf{Adversarial Loss.}}
To enforce the translation networks to produce data that is indistinguishable from genuine images, we also include an adversarial loss to train our model. For both translators, in GAN context often called generators, we train two additional discriminator networks $D_A$ and $D_B$ that are trained to distinguish translated from genuine images. To train our model, we use the following adversarial term:
\begin{equation}
\mathcal{L}_{adv} = \mathcal{L}_{GAN}(G_A, D_A) + \mathcal{L}_{GAN}(G_B, D_B),
\end{equation}
where $\mathcal{L}_{GAN}(G, D)$ is the LSGAN formulation~\cite{mao2017least}.

Overall, we train our model using the following loss function:
\begin{equation}
\label{equation:Total}
\mathcal{L} = \lambda_{adv}\mathcal{L}_{adv} + \lambda_{cyc}\mathcal{L}_{cyc} + \lambda_{corr}\mathcal{L}_{corr},
\end{equation}
where $\lambda_{adv}$, $\lambda_{cyc}$, and $\lambda_{corr}$ are user-defined hyperparameters.

\subsection{Implementation Details}
\label{sec:impl_details}
\begin{figure*}[t!]
\centering
{\includegraphics[width=.99\linewidth]{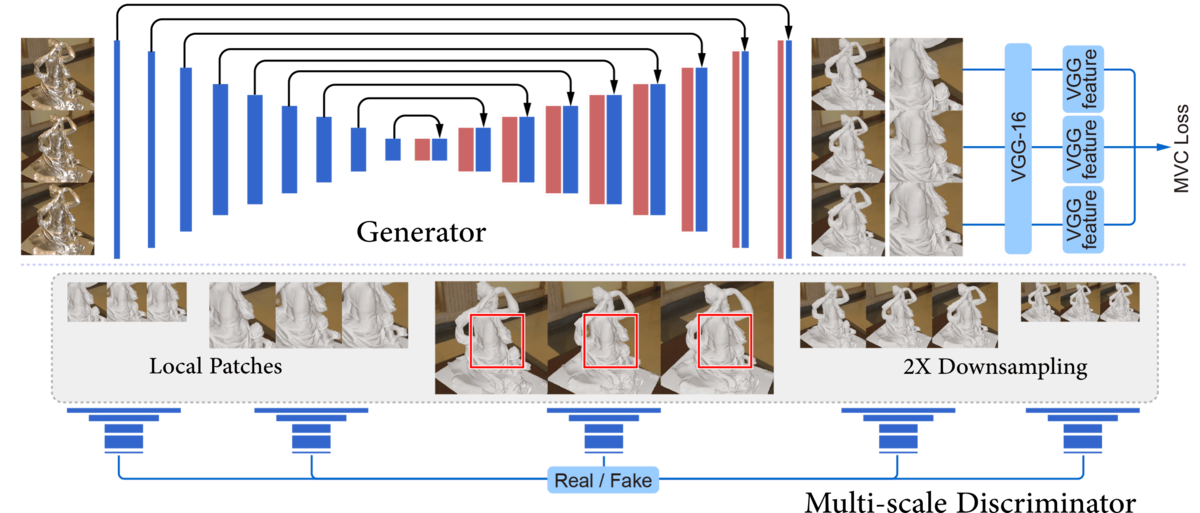}}
\caption{Illustration of the generator and discriminator network. The generator uses the U-net architecture and both input and output are a multi-view sequence consisting of three views. A random SIFT correspondence is sampled during training to compute the correspondence loss. The multi-scale joint discriminator examines three scales of the image sequence and two scales of corresponding local patches. The width and height of each rectangular block indicate the channel size and the spatial dimension of the output feature map, respectively.}
\label{fig:network}
\end{figure*}
Our model is based on cycleGAN and implemented in Pytorch. We experimented with different architectures for the translation networks, including U-Net \cite{ronneberger2015u}, ResNet \cite{He_2016_CVPR}, and RNN-blocks \cite{chaitanya2017interactive}. Given enough training time, we found that all networks produce similar results. Due to its memory efficiency and fast convergence, we chose U-Net for our final model. As shown in Figure~\ref{fig:network}, we use the multi-scale discriminator introduced in \cite{wang2017high} that downsamples by a rate of 2, which generally works better for high resolution images. Our discriminator also considers the local correspondence patches as additional input, which helps to produce coherent translations. Followed by the training guidances proposed in \cite{karras2017progressive}, we use pixel-wise normalization in the generators and add a 1-strided convolutional layer after each deconvolutional layer. For computing the correspondence loss, we use a patch size of $256\times256$ and sample a single SIFT correspondence per training iteration randomly. The discriminator follows the architecture as: C64-C128-C256-C512-C1. The generator's encoder architecture is: C64-C128-C256-C512-C512-C512-C512-C512. We use $\lambda_{adv} = 1, \lambda_{cyc} = 10, \lambda_{corr} = 5$ in all our experiments and train using the ADAM optimizer with a learning rate of 0.0002.

\section{Evaluation}
\label{sec:results}

In this section, we present qualitative and quantitative evaluations of our proposed S2Dnet. For this purpose, we evaluate the performance of our model on both the translation task and the subsequent 3D reconstruction, and we compare to several baseline systems. In Section~\ref{sec:synthetic_results} we report results on our synthetic test set and we also perform an evaluation on real-world data in Section~\ref{sec:real_results}.

To evaluate the benefit of our proposed inter-view coherence, we perform a comparison to a single-view translation baseline by training a cycleGAN network~\cite{zhu2017unpaired} on glossy to diffuse translation. Since our synthetic dataset features a bijective mapping between glossy and diffuse images, we also train a pix2pix network~\cite{isola2017image} for a supervised baseline on synthetic data. In addition, we compare reconstruction quality to performing stereo reconstruction directly on the glossy multi-view sequence to demonstrate the benefit of translating the input as a preprocessing step. For 3D reconstruction, we apply a state-of-the-art multi-view surface reconstruction method~\cite{Langguth2016} on input sequences consisting of 10 to 15 views. For our method, we translate each input view sequentially but we feed the two neighboring views as additional inputs to our multi-view network. For the two baseline translation methods, we translate each view independently. The 3D reconstruction pipeline then uses the entire translated multi-view sequence as input.

\subsection{Synthetic Data}
\label{sec:synthetic_results}

\begin{table}[t!]
\centering
\begin{tabular}{||c|c|c|c|c||} \hline
                                  & Glossy  &  pix2pix   &  cycleGAN  &  S2Dnet 
\\ \hline Image MSE		            & 118.39  &      56.20 &      69.15 & 57.78  
\\ \hline
\end{tabular}
\caption{Quantitative evaluation of the image error on our synthetic testing data.}
\label{tab:pixel_error}
\end{table}

\begin{figure*}[t!]
\centering
{\includegraphics[width=.9\linewidth]{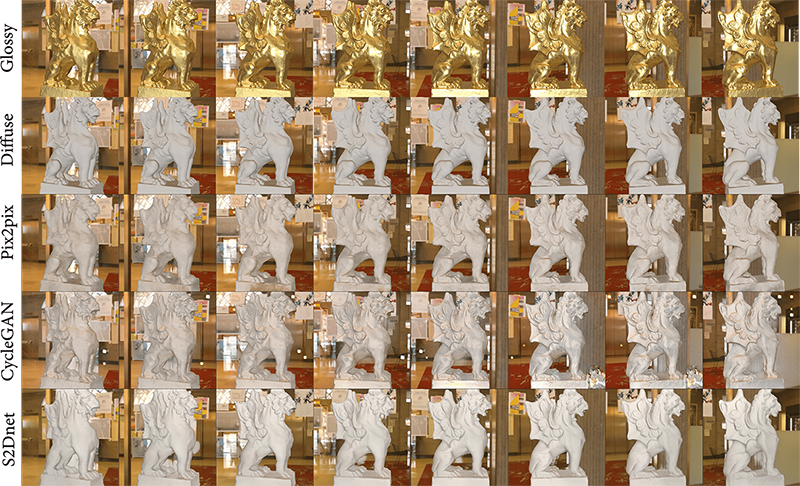}}   
\caption{Qualitative translation results on a synthetic input sequence consisting of 8 views. From top down: the glossy input sequence, the ground truth diffuse rendering, and the translation results for the baselines pix2pix and cycleGAN, and our S2Dnet. The output of pix2pix is generally blurry. The cycleGAN output, although sharp, lacks inter-view consistency. Our S2Dnet produces both crisp and coherent translations.}
\label{fig:qualitative_synthetic_translation}
\end{figure*}
For a quantitative evaluation of the image translation performance, we compute MSE with respect to the ground truth diffuse renderings on our synthetic test set. Table~\ref{tab:pixel_error} shows a comparison of our S2Dnet to pix2pix and cycleGAN. Unsurprisingly, the supervised pix2pix network performs best, closely followed by our S2Dnet, which outperforms the unsupervised baseline by a significant margin. In Figure~\ref{fig:qualitative_synthetic_translation} we show qualitative translation results. Note that the output of pix2pix is generally blurry. Since MSE penalizes outliers and prefers a smooth solution, pix2pix still achieves a low MSE error. While the output of cycleGAN is sharper, the translated sequence lacks inter-view consistency, whereas our S2Dnet produces both highly detailed and coherent translations.

\begin{table}[t!]
\centering
\begin{tabular}{||c|c|c|c|c|c|c|c|c|c|c|c||} \hline
              Model           & 1  &  2   &  3  &  4 &  5 &  6 &  7 &  8 &  9 &  10   & AVG  
\\ \hline Glossy		          &  0.67	& 0.88 & 1.35 & 0.76 & 1.15 & 1.13 & 1.15 & 0.78 & 0.54 & 0.66 & 0.90    
\\ \hline cycleGAN            &  1.18	& 0.72 & 0.89 & 0.59 & 1.35 & 0.72 & 0.99 & 0.62 & 0.51 & 0.42 & 0.80 
\\ \hline S2Dnet              &  0.52	& 0.67 & 0.72 & 0.43 & 0.87 & 0.54 & 0.92 & 0.65 & 0.55 & 0.56 & 0.64 
\\ \hline
\end{tabular}
\caption{Quantitative evaluation of surface reconstruction performance on 10 different scenes. The error metric is the percentage of bounding box diagonal.  Our S2Dnet performs best, and the translation baseline still performs significantly better than directly reconstructing from the glossy images. The numbering of the models follows the visualization in Figure~\ref{fig:qualitative_synthetic_reconstruction}, using the same left to right order.}
\label{tab:geometry_error}
\end{table}

\begin{figure*}[t!]
\centering
{\includegraphics[width=.9\linewidth]{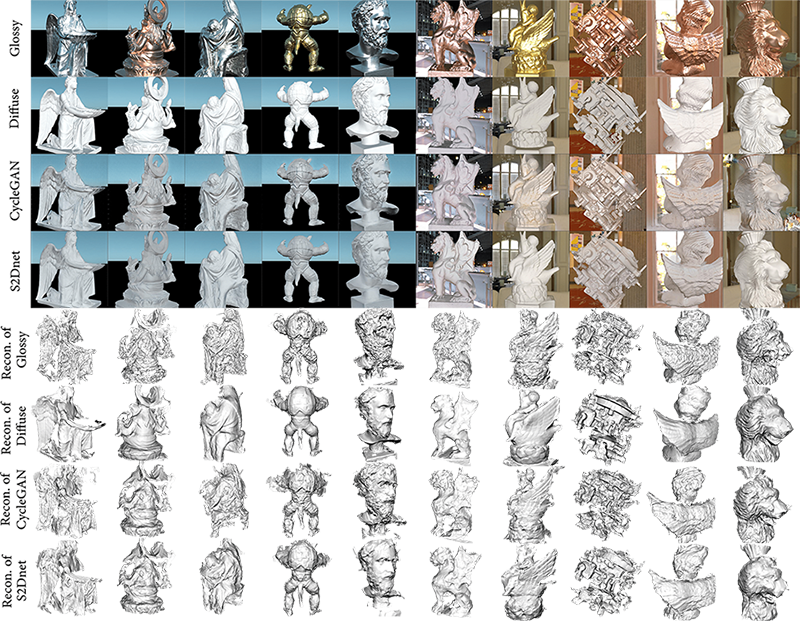}}	
\caption{Qualitative surface reconstruction results on 10 different scenes. From top to bottom: glossy input, ground truth diffuse renderings, cycleGAN translation outputs, our S2Dnet translation outputs, reconstructions from glossy images, reconstructions from ground truth diffuse images, reconstructions from cycleGAN output, and reconstructions from our S2Dnet output. All sequences are excluded from our training set, and the objects in column 3 and 4 have not even been seen during training.}
\label{fig:qualitative_synthetic_reconstruction}
\end{figure*}
Next, we evaluate the quality of the surface reconstruction by feeding the translated sequences to the reconstruction pipeline. We found that the blurry output of pix2pix is not suitable for stereo reconstruction, since already the first step, estimating camera parameters based on feature correspondences, fails on this data. We therefore exclude pix2pix from the surface reconstruction evaluation but include the trivial baseline of directly reconstructing from the glossy input sequence to demonstrate the benefit of the translation step. In order to compute the geometric error of the surface reconstruction output, we register the reconstructed geometry to the ground truth mesh using a variant of ICP \cite{rusinkiewicz2001efficient}. Next, we compute the Euclidean distance of each reconstructed surface point to its nearest neighbor in the ground truth mesh and report the per-model mean value. Table~\ref{tab:geometry_error} shows the surface reconstruction error for our S2Dnet in comparison to the three baselines. The numbers show that our S2Dnet performs best, and that preprocessing the glossy input sequences clearly helps to obtain a more accurate reconstruction, even when using the cycleGAN baseline. In Figure~\ref{fig:qualitative_synthetic_reconstruction} we show qualitative surface reconstruction results for 10 different scenes in various environments. 

\subsection{Real-world Data}
\label{sec:real_results}
Since we do not have real-world ground truth data, we compile a real-world test set and perform a qualitative comparison on it. For all methods, we compare generalization performance when training on our synthetic dataset. Moreover, we evaluate how the different models perform when fine-tuning on real-world data, or training on real-world data from scratch. For this purpose, we compile a dataset by shooting photos of real-world objects. We choose 5 diffuse real-world objects and take 5k pictures in total from different camera positions and under varying lighting conditions. Next, we use a glossy spray paint to cover our objects with a glossy coat and shoot another 5k pictures to represent the glossy domain. The resulting dataset consists of unpaired samples of glossy and diffuse objects under real-world conditions, see Figure~\ref{fig_domain} a) and b).

\begin{figure*}[t!]
\centering
{\includegraphics[width=.85\linewidth]{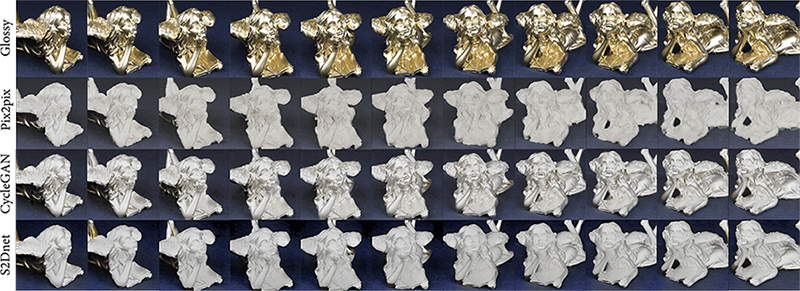}}
\caption{Qualitative translation results on a real-world input sequence consisting of 11 views. The first row shows the glossy input sequence and the remaining rows show the translation results of pix2pix, cycleGAN, and our S2Dnet. All networks are trained on synthetic data only. Similar to the synthetic case, cycleGAN outperforms pix2pix, but it produces high-frequency artifacts that are not consistent along the views. Our S2Dnet is able to remove most of the specular effects and preserves all the geometric details in a consistent manner.}
\label{fig:qualitative_real_translation}
\end{figure*}
\begin{figure*}[t!]
\centering
{\includegraphics[width=.85\linewidth]{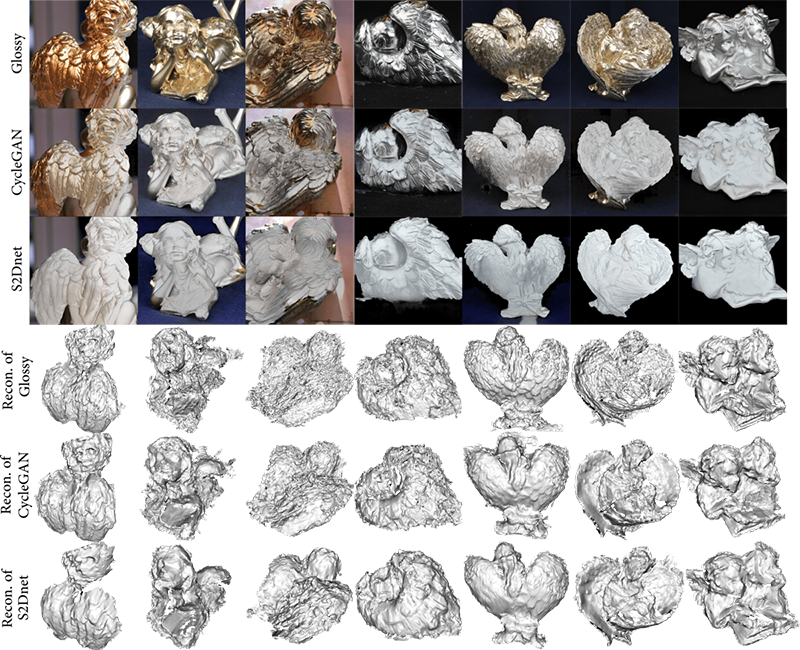}}
\caption{Qualitative surface reconstruction results on 7 different real-world scenes. Top to bottom: glossy input, cycleGAN translation outputs, our S2Dnet translation outputs, reconstructions from glossy images, reconstructions from cycleGAN output, and reconstructions from our S2Dnet output. All networks are trained on synthetic data only.}
\label{fig:qualitative_real_reconstruction}
\end{figure*}
In Figure~\ref{fig:qualitative_real_translation} we show qualitative translation results on real-world data. All networks are trained on synthetic data only here, and they all manage to generalize to some extent to real-world data, thanks to our high-quality synthetic dataset. Similar to the synthetic results in Figure~\ref{fig:qualitative_synthetic_translation}, pix2pix produces blurry results, while cycleGAN introduces inconsistent high-frequency artifacts. S2Dnet is able the remove most of the specular effects and preserves geometric details in a consistent manner. In Figure~\ref{fig:qualitative_real_reconstruction} we show qualitative surface reconstruction results for 7 different scenes. Artifacts occur mainly close to the object silhouettes in complex background environments. This could be mitigated by training with segmentation masks.

\begin{figure*}[t!]
\centering
	\subfloat[]
	{\includegraphics[height=.15\linewidth]{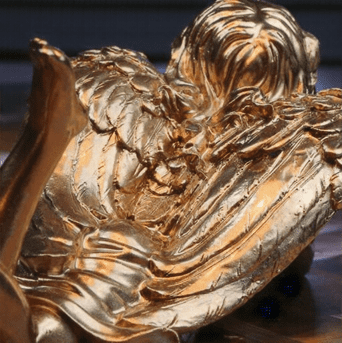}}  
	\subfloat[]
	{\includegraphics[height=.15\linewidth]{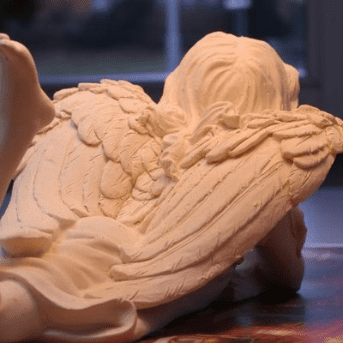}} 
	\subfloat[]
	{\includegraphics[height=.15\linewidth]{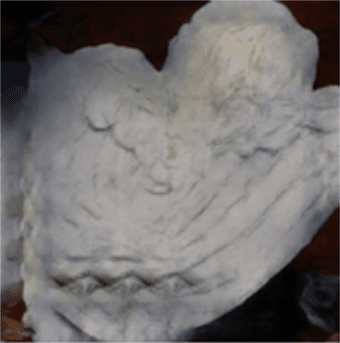}} 
	\subfloat[]
	{\includegraphics[height=.15\linewidth]{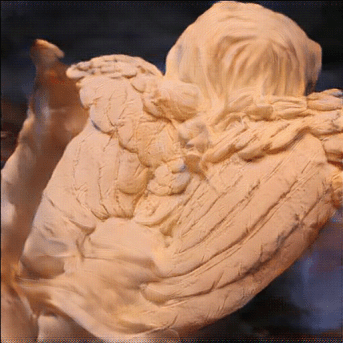}}   
	\subfloat[]
	{\includegraphics[height=.15\linewidth]{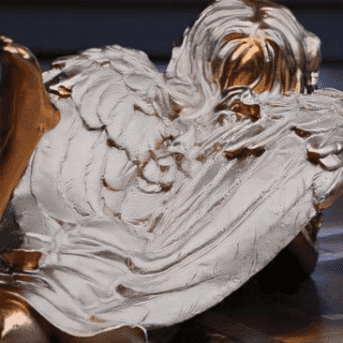}}  
	\subfloat[]
	{\includegraphics[height=.15\linewidth]{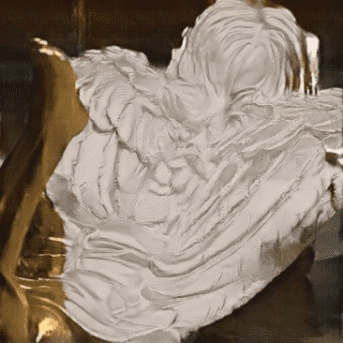}}  
	
\caption{a), b) A sample of our real-world dataset. c) translation result of cycleGAN when training from scratch on our real-world dataset. d) S2Dnet output, trained from scratch on our real-world dataset. e) S2Dnet output, trained on synthetic data only. f) S2Dnet output, trained on synthetic data, fine-tuned on real-world data.}
\label{fig_domain}
\end{figure*}

Finally, we evaluate performance when either fine-tuning or training from scratch on real-world data. We retrain or fine-tune S2Dnet and cycleGAN on our real-world dataset, but cannot retrain pix2pix for this purpose, since it relies on a supervision signal that is not present in our unpaired real-world dataset. Our experiments show that training or fine-tuning using such a small dataset leads to heavy overfitting. The translation performance for real-world objects that were not seen during training decreases significantly compared to the models trained on synthetic data only. In Figure~\ref{fig_domain} c) and d) we show image translation results of cycleGAN and S2Dnet when training from scratch on our real-world dataset. Since the scene in Figure~\ref{fig_domain} is part of the training set (although the input image itself is excluded from the training set), our S2Dnet produces decent translation results, which is not the case for scenes not seen during training. Fine-tuning our S2Dnet produces similar results (Figure~\ref{fig_domain} f)).

\section{Limitations and Future Work}
\label{sec:future}

Although the proposed framework enables reconstructing glossy and specular objects more accurately compared to state-of-the-art 3D reconstruction algorithms, a few limitations do exist.
First, since the network architecture contains an encoder and a decoder with skip connections, the glossy-to-Lambertian image translation is limited to images of a fixed resolution.
This resolutions might be too low for certain types of applications.
Next, due to the variability of the background in real images, the translation network might treat a portion of the background as part of the reconstructed object.
Similarly, the network occasionally misclassifies the foreground as part of the background, especially in very light domains on specular objects.
Finally, as the simulated training data was rendered by assuming a fixed albedo, the network cannot consistently translate glossy materials with spatially varying albedo into a Lambertian surface. 
We predict that given a larger and more diverse training set in terms of shapes, backgrounds, albedos and materials, the accuracy of the proposed method in recovering real object would be largely enhanced. Our current training dataset includes the most common types of specular material. The proposed translation network has potential to be extended to other more challenging materials, such as transparent objects, given proper training data.

\section{Supplementary Material}

This supplementary material provides more technical details and experimental results for our specular-to-diffuse translation network, S2Dnet. Upon publication we will also release the full training and test data set as well as the trained network and code.

\paragraph{\bf{Training Details.}}

Tables~\ref{tab:discri} and~\ref{tab:generator} give more detail of our network architecture. Xavier \cite{glorot2010understanding} is used for weights initialization. We train our models on an NVIDIA 1080Ti GPU with 11GB GPU memory, which only allows us to use a training batch size of 1.

\paragraph{\bf{Additional Results.}}

\begin{figure*}[htbp!]
\centering
{\includegraphics[width=1.0\linewidth]{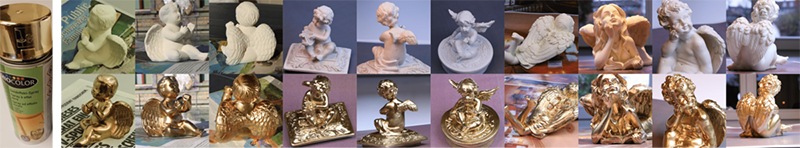}}	
\caption{Gallery of our glossy-to-diffuse real-world training data and the spray (leftmost column) we used to paint the objects. We first choose 5 diffuse real-world objects and take 5k pictures in total from different camera positions and under varying lighting conditions. We then use a glossy spray paint to cover our objects with a glossy coat and shoot another 5k pictures to represent the glossy domain.}
\label{fig:real_train}
\end{figure*}

\begin{figure*}[htbp!]
\centering
{\includegraphics[width=1.0\linewidth]{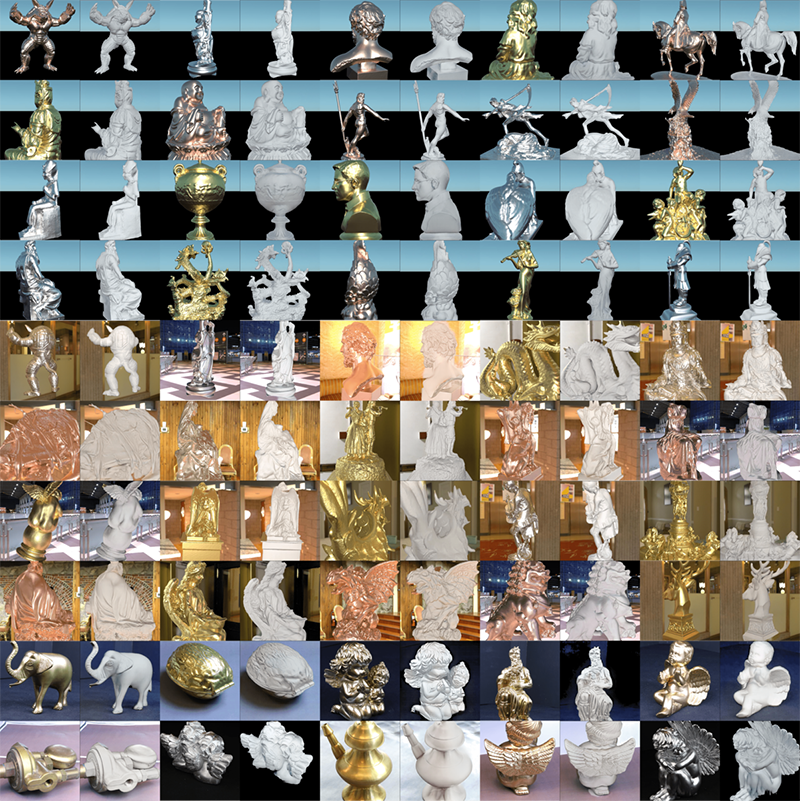}}	
\caption{Gallery of our glossy-to-diffuse results of 40 synthetic and 10 real-world (last two rows) scenes. All sequences are excluded from our training set. Three synthetic (Armadillo, Standing Buddha, Roman Head Sculpture) and all real-world objects have not even been seen during training.}
\label{fig:results}
\end{figure*}

\begin{figure*}[htbp!]
\centering
{\includegraphics[width=1.0\linewidth]{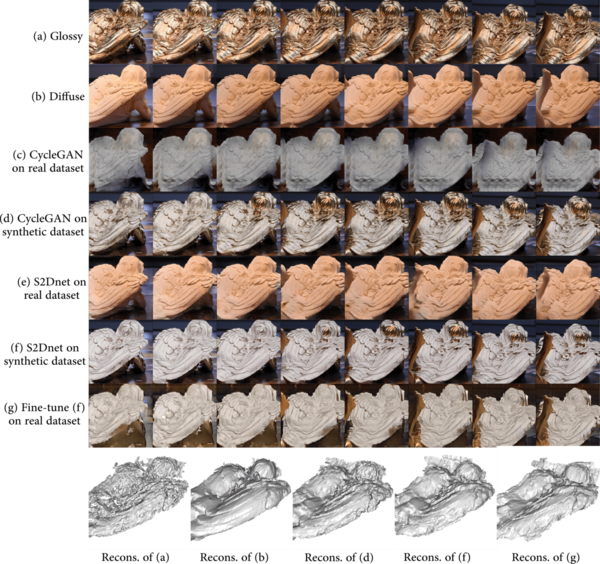}}	
\caption{A sample  of our real-world dataset is shown in (a-b). Translation results of cycleGAN when training from scratch on our real-world dataset or synthetic data only are shown in (c) and (d), respectively. S2Dnet outputs, trained from scratch on our real-world dataset or synthetic data only, are shown in (e) and (f), respectively. Another output of S2Dnet, trained on synthetic data and then fine-tuned on real-world data is presented in (g). The last row demonstrates the corresponding reconstruction results. Note that the output images are blurry when training from scratch on real-world data, i.e. (c) and (e), and thus not suitable for stereo reconstruction.}
\label{fig:real_finetune}
\end{figure*}

\begin{figure*}[htbp!]
\centering
{\includegraphics[width=1.0\linewidth]{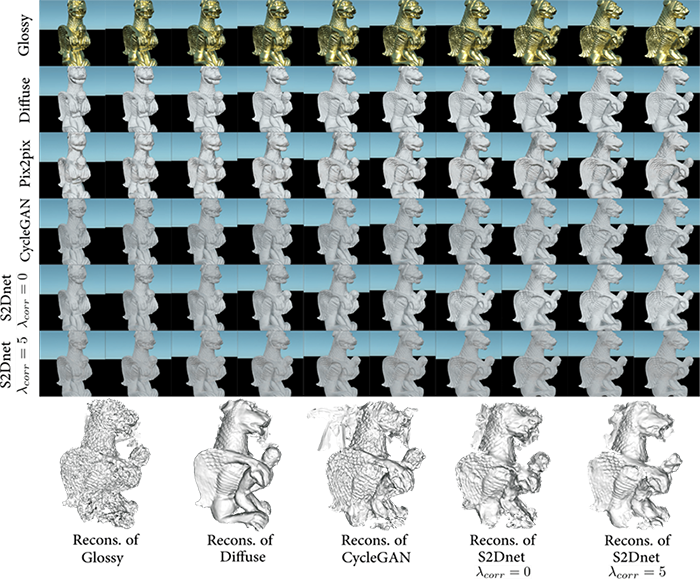}}	
\caption{Qualitative comparison of image translation and surface reconstruction on a synthetic sequence consisting of 10 views. From top to bottom: glossy input, ground truth diffuse renderings, pix2pix translation outputs, cycleGAN translation outputs, our S2Dnet translation outputs using $\lambda_{corr} = 0$ (no perceptual correspondence loss), our S2Dnet translation outputs using $\lambda_{corr} = 5$. The last row shows the corresponding reconstruction results. All sequences are excluded from our training set. The output of pix2pix is blurry and is not suitable for multi-view reconstruction. The outputs of cycleGAN and our S2Dnet without perceptual correspondence loss, although sharp, lack of inter-view consistency. Our S2Dnet with perceptual correspondence loss ($\lambda_{corr} = 5$) produces crisp and coherent translations, resulting in a better reconstruction.}
\label{fig:lambda1}
\end{figure*}

\begin{figure*}[htbp!]
\centering
{\includegraphics[width=0.82\linewidth]{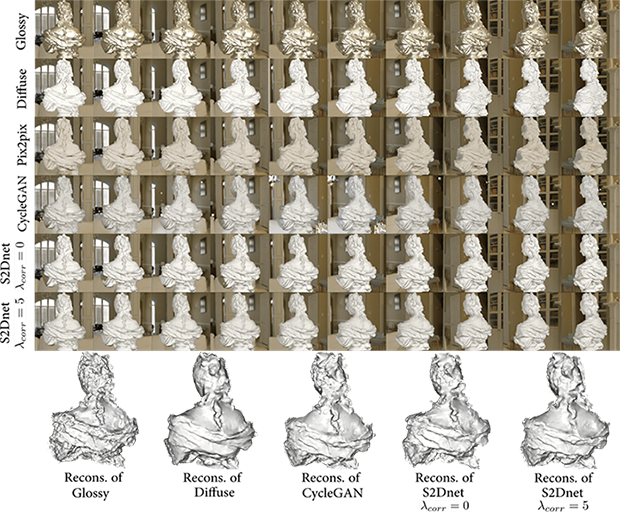}}	
\caption{Another set of image translation and surface reconstruction comparison on a synthetic input sequence consisting of 10 views.}
\label{fig:lambda2}
\end{figure*}

Figure~\ref{fig:real_train} presents a gallery of our real-world training data as described in the paper. Figure~\ref{fig:results} shows more results of our S2Dnet, given input scenes with various illumination and different objects. Figure~\ref{fig:real_finetune} is an extension of Figure 10 in paper, illustrating the results of training setups using different kinds of training data. Figures~\ref{fig:lambda1} and~\ref{fig:lambda2} demonstrate two additional examples of image translation and reconstruction, where S2Dnet outperforms both pix2pix and cycleGAN. We also observe that when the weight for the perceptual correspondence loss is $\lambda_{corr} = 0$, i.e., removing the perceptual correspondence loss, the output of S2Dnet lacks of inter-view consistency. 


\begin{table}[h!]
\centering
\begin{tabular}{c c c} 

	Layer & Input $\rightarrow$ Output Shape & Layer Information\\
	\hline \hline
	
	Input Layer   & (h, 3w, 3) $\rightarrow$ ($\frac{h}{2}$, $\frac{3w}{2}$, 64) & CONV-(N64, K4x4, S2, P1), LeReLU\\

	 \hline
	
	\multirow{3}{*}{Hidden Layers}
								& ($\frac{h}{2}$, $\frac{3w}{2}$, 64) $\rightarrow$ ($\frac{h}{4}$, $\frac{3w}{4}$, 128) & CONV-(N128, K4x4, S2, P1), PN, LeReLU\\
								& ($\frac{h}{4}$, $\frac{3w}{4}$, 128) $\rightarrow$ ($\frac{h}{8}$, $\frac{3w}{8}$, 256) & CONV-(N256, K4x4, S2, P1), PN, LeReLU\\
								& ($\frac{h}{8}$, $\frac{3w}{8}$, 256) $\rightarrow$ ($\frac{h}{16}$, $\frac{3w}{16}$, 512) & CONV-(N512, K4x4, S2, P1), PN, LeReLU\\
								
		\hline
									
	Output Layer  & ($\frac{h}{16}$, $\frac{3w}{16}$, 512) $\rightarrow$ ($\frac{h}{32}$, $\frac{3w}{32}$, 1) & CONV-(N1, K4x4, S2, P1)\\	
																																				
			\hline \hline																																																																																																																											
\end{tabular}

\caption{Discriminator network architecture. We use 5 such discriminators that have an identical network structure but operate at three scales of the image sequence and two scales of corresponding local patches using LSGAN (see Figure 5 in the paper). N: the number of output channels, K: kernel size, S: stride size, P: padding size, PN: pixel-wise Normalization, LeReLU: LeakyReLU with $\alpha=0.2$, $w, h$: width and height of input images. Note that the input width is $3w$ because we spatially concatenate the three views of the input sequences.}
\label{tab:discri}
\end{table}

\begin{table}[htbp!]
\centering
\begin{tabular}{c c c} 

	Part & Input $\rightarrow$ Output Shape & Layer Information\\
	\hline \hline
	
	\multirow{9}{*}{Down-sampling}
								& (h, 3w, 3) $\rightarrow$ ($\frac{h}{2}$, $\frac{3w}{2}$, 64) & CONV-(N64, K4x4, S2, P1), LeReLU\\
								& ($\frac{h}{2}$, $\frac{3w}{2}$, 64) $\rightarrow$ ($\frac{h}{4}$, $\frac{3w}{4}$, 128) & CONV-(N128, K4x4, S2, P1), PN, LeReLU\\
								& ($\frac{h}{4}$, $\frac{3w}{4}$, 128) $\rightarrow$ ($\frac{h}{8}$, $\frac{3w}{8}$, 256) & CONV-(N256, K4x4, S2, P1), PN, LeReLU\\
								& ($\frac{h}{8}$, $\frac{3w}{8}$, 256) $\rightarrow$ ($\frac{h}{16}$, $\frac{3w}{16}$, 512) & CONV-(N512, K4x4, S2, P1), PN, LeReLU\\
								& ($\frac{h}{16}$, $\frac{3w}{16}$, 512) $\rightarrow$ ($\frac{h}{32}$, $\frac{3w}{32}$, 512) & CONV-(N512, K4x4, S2, P1), PN, LeReLU\\
								& ($\frac{h}{32}$, $\frac{3w}{32}$, 512) $\rightarrow$ ($\frac{h}{64}$, $\frac{3w}{64}$, 512) & CONV-(N512, K4x4, S2, P1), PN, LeReLU\\
								& ($\frac{h}{64}$, $\frac{3w}{64}$, 512) $\rightarrow$ ($\frac{h}{128}$, $\frac{3w}{128}$, 512) & CONV-(N512, K4x4, S2, P1), PN, LeReLU\\
								& ($\frac{h}{128}$, $\frac{3w}{128}$, 512) $\rightarrow$ ($\frac{h}{256}$, $\frac{3w}{256}$, 512) & CONV-(N512, K4x4, S2, P1), PN, LeReLU\\
								& ($\frac{h}{256}$, $\frac{3w}{256}$, 512) $\rightarrow$ ($\frac{h}{512}$, $\frac{3w}{512}$, 512) & CONV-(N512, K4x4, S2, P1), ReLU\\
								
		\hline
	  \multirow{18}{*}{Up-sampling}
		
								& \multirow{2}{*}{ ($\frac{h}{512}$, $\frac{3w}{512}$, 512) $\rightarrow$ ($\frac{h}{256}$, $\frac{3w}{256}$, 512) } & DECONV-(N512, K4x4, S2, P1), \\
								&                                                                                                                    &    CONV-(N512, K3x3, S1, P1), PN, ReLU\\
								
								& \multirow{2}{*}{ ($\frac{h}{256}$, $\frac{3w}{256}$, 1024) $\rightarrow$ ($\frac{h}{128}$, $\frac{3w}{128}$, 512) } & DECONV-(N512, K4x4, S2, P1), \\
								&                                                                                                                   &    CONV-(N512, K3x3, S1, P1), PN, ReLU\\
								
								& \multirow{2}{*}{ ($\frac{h}{128}$, $\frac{3w}{128}$, 1024) $\rightarrow$ ($\frac{h}{64}$, $\frac{3w}{64}$, 512) } & DECONV-(N512, K4x4, S2, P1), \\
								&                                                                                                                   &    CONV-(N512, K3x3, S1, P1), PN, ReLU\\
														
								& \multirow{2}{*}{ ($\frac{h}{64}$, $\frac{3w}{64}$, 1024) $\rightarrow$ ($\frac{h}{32}$, $\frac{3w}{32}$, 512) } & DECONV-(N512, K4x4, S2, P1), \\
								&                                                                                                                   &    CONV-(N512, K3x3, S1, P1), PN, ReLU\\
								
								& \multirow{2}{*}{ ($\frac{h}{32}$, $\frac{3w}{32}$, 1024) $\rightarrow$ ($\frac{h}{16}$, $\frac{3w}{16}$, 512) } & DECONV-(N512, K4x4, S2, P1), \\
								&                                                                                                                   &    CONV-(N512, K3x3, S1, P1), PN, ReLU\\
																								
								& \multirow{2}{*}{ ($\frac{h}{16}$, $\frac{3w}{16}$, 1024) $\rightarrow$ ($\frac{h}{8}$, $\frac{3w}{8}$, 256) } & DECONV-(N256, K4x4, S2, P1), \\
								&                                                                                                                   &    CONV-(N256, K3x3, S1, P1), PN, ReLU\\
																																
								& \multirow{2}{*}{ ($\frac{h}{8}$, $\frac{3w}{8}$, 512) $\rightarrow$ ($\frac{h}{4}$, $\frac{3w}{2}$, 128) } & DECONV-(N128, K4x4, S2, P1), \\
								&                                                                                                                   &    CONV-(N128, K3x3, S1, P1), PN, ReLU\\
																																								
								& \multirow{2}{*}{ ($\frac{h}{4}$, $\frac{3w}{4}$, 256) $\rightarrow$ ($\frac{h}{2}$, $\frac{3w}{2}$, 64) } & DECONV-(N64, K4x4, S2, P1), \\
								&                                                                                                                   &    CONV-(N64, K3x3, S1, P1), PN, ReLU\\																																								
								& \multirow{2}{*}{ ($\frac{h}{2}$, $\frac{3w}{2}$, 64)  $\rightarrow$ (h, 3w, 3) } & DECONV-(N3, K4x4, S2, P1), \\
								&                                                                                                                   &    CONV-(N512, K3x3, S1, P1), Tanh\\																																								
			\hline \hline																																					
																																																																																																																																										
\end{tabular}

\caption{Generator network architecture. }
\label{tab:generator}
\end{table}

\section*{Acknowledgement}
We thank the anonymous reviewers for their constructive comments. This work was supported in parts by Swiss National  Science Foundation (169151), NSFC (61522213, 61761146002, 61861130365), 973 Program (2015CB352501), Guangdong Science and Technology Program (2015A030312015), ISF-NSFC Joint Research Program (2472/17) and Shenzhen Innovation Program (KQJSCX20170727101233642).

\clearpage

\bibliographystyle{splncs}
\bibliography{MVR}
\end{document}